\documentclass[runningheads]{llncs}
\usepackage[T1]{fontenc}
\usepackage{graphicx}
\usepackage{amsmath}
\usepackage{algorithm}
\usepackage{algorithmic}

\begin{document}
\title{Breaking the weakest link to evade vision language models}
\author{Ilan Zini\inst{1} \and
Boussad Addad\inst{2} \and
Katarzyna Kapusta\inst{2}}
\authorrunning{I. Zini et al.}
%

\institute{ESILV student and Thales intern \\
\email{ilanzini.pro@gmail.com}\\ \and
Thales, Thales cortAIx Labs, France\\
\email{\{boussad.addad, katarzyna.kapusta\}@thalesgroup.com}}
\maketitle 

\begin{abstract}
Vision--Language Models (VLMs) have recently emerged as a critical component of multimodal AI systems, enabling joint reasoning over visual and textual inputs in real-world and safety-critical applications. Despite their growing deployment, the robustness of VLMs against adversarial threats remains insufficiently explored, particularly in the context of evasion attacks targeting multimodal alignment. In this work, we investigate the vulnerability of VLMs to adversarial perturbations applied to visual inputs and study two attack settings: untargeted attacks, where the goal is to disrupt the model's interpretation of the original image, and targeted attacks, where the adversary aims to force the model to generate a specific semantic description unrelated to the original image. To efficiently generate adversarial examples, we propose a gradient-based attack method that performs optimization exclusively on the vision encoder of the VLM rather than on the entire multimodal architecture. This design significantly reduces the computational cost and resource requirements of the attack while maintaining strong effectiveness. We evaluate our approach on several open-source VLMs, including Qwen2.5-VL, Granite-Vision, FastVLM, and Phi-3.5-Vision, and show that small, human-imperceptible perturbations can substantially alter the textual interpretation produced by the models. Our findings highlight the vulnerability of modern VLMs to adversarial manipulation and emphasize the need for improved robustness and security mechanisms in multimodal AI systems.

\end{abstract}

\begin{figure}
\includegraphics[width=\textwidth]{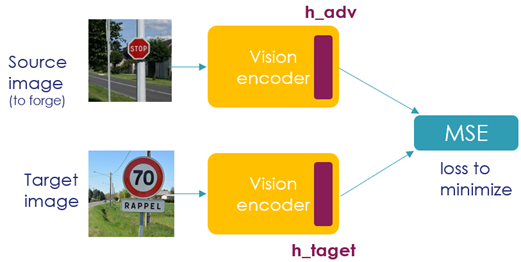}
\caption{Targeted adversarial attack using vision encoder embedding alignment.}\label{fig1}
\end{figure}

\begin{figure}
\includegraphics[width=\textwidth]{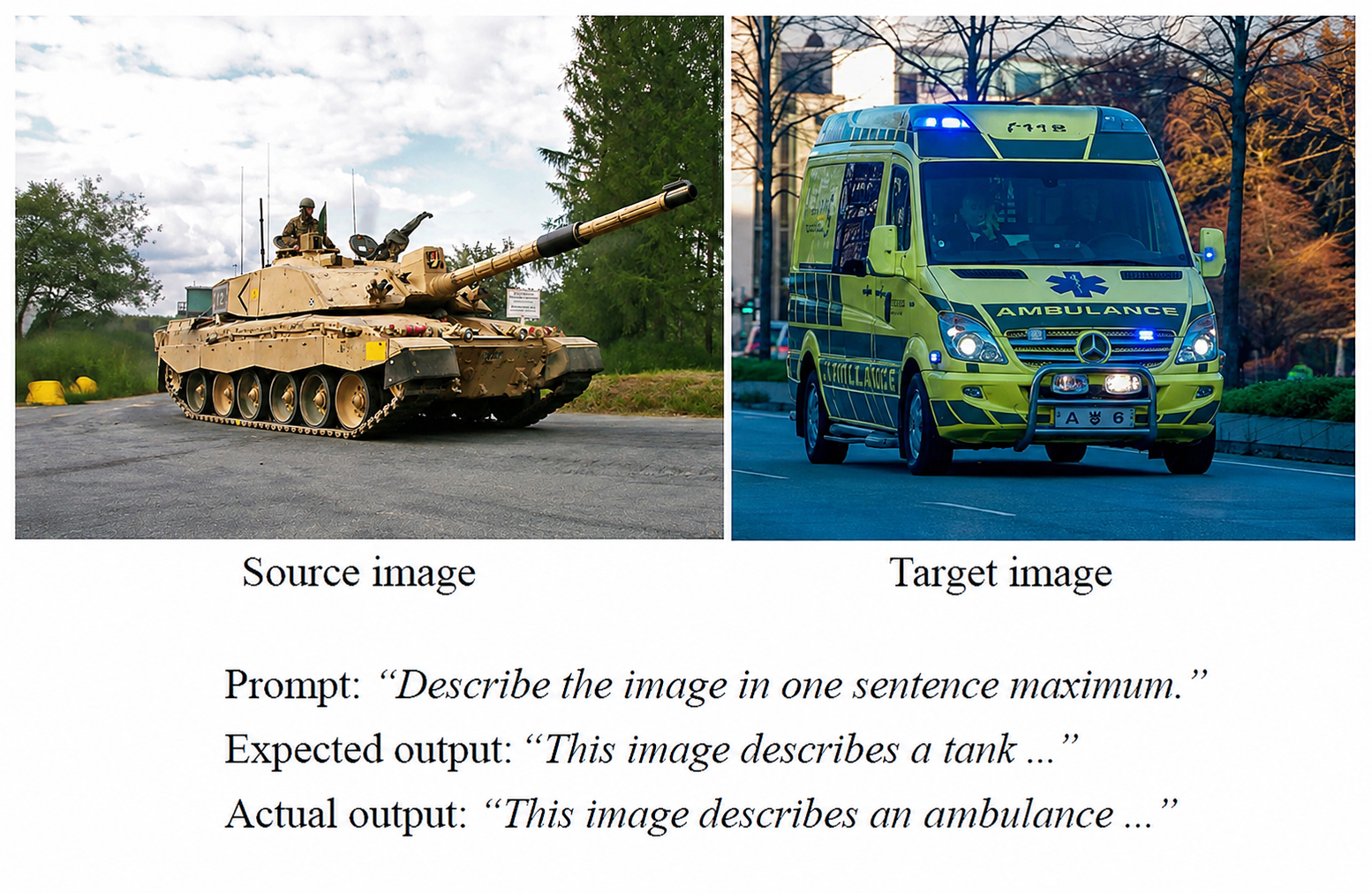}
\caption{Targeted adversarial attack forcing a VLM to misinterpret a military tank as an ambulance.}\label{fig2}
\end{figure}

\section{Introduction}

Multi-modal foundation models that combine vision and language have recently attracted significant attention. By integrating powerful large language models with visual encoders, these systems are capable of jointly processing images and text, enabling a wide range of applications such as image captioning, visual question answering, and multimodal reasoning. In these tasks, the model must extract meaningful visual representations from images and combine them with linguistic context to produce coherent textual outputs. As a result, Vision--Language Models (VLMs) have become a key component in modern AI systems deployed in real-world applications.

Despite their impressive capabilities, the increasing deployment of multimodal models also raises important security concerns. Models operating in open environments can be exposed to adversarial manipulation, where attackers deliberately modify inputs to alter the behavior of the system. In the visual domain, adversarial examples have long been known to exploit the vulnerability of neural networks by introducing small perturbations to images that remain imperceptible to humans but can drastically change the model's predictions~\cite{szegedy2013,biggio2013}. When applied to vision--language models, such perturbations can cause the system to generate incorrect or misleading textual descriptions of visual content.

These vulnerabilities pose serious risks in practical scenarios. Malicious actors could manipulate images to spread misinformation, bias model outputs, or generate harmful content while maintaining the appearance of legitimate model responses. Because adversarial perturbations are typically very small and difficult for humans to detect, users may unknowingly trust manipulated outputs generated by the model. Figure~\ref{fig2} illustrates such a scenario, where a targeted adversarial attack forces a VLM to describe a military tank image as an ambulance. Understanding how such attacks can be constructed and how they affect multimodal models is therefore essential for assessing the robustness and safety of these systems.

In this work, we study adversarial attacks against vision--language models and investigate how small perturbations applied to input images can alter the textual outputs generated by the model. We focus on two attack settings: targeted attacks, where the adversary aims to force the model to produce a specific output, and untargeted attacks, where the goal is simply to disrupt the model's interpretation of the original image.

Unlike previous work that performs adversarial optimization through the entire multimodal architecture, we propose a more efficient approach that focuses only on the vision encoder component of the VLM. We consider a white-box threat model in which the attacker has full access to the vision encoder. This assumption is realistic in practice, as many competitive VLMs are publicly released as open-source systems. An attacker can therefore craft adversarial examples against a known open-source model and deploy them against any system that relies on the same architecture or vision encoder. More specifically, by performing backpropagation exclusively through the visual encoder, we are able to generate effective adversarial examples while significantly reducing the computational cost of the attack. This approach enables faster and more resource-efficient generation of adversarial images while maintaining strong attack performance across multiple open-source VLM models. An overview of this approach is illustrated in Figure~\ref{fig1}.

\section{Background and related work}

Adversarial attacks against machine learning models have been extensively studied over the past decade, with two main threat categories: evasion attacks, where the adversary modifies inputs at test time, and poisoning attacks, where the training data is corrupted~\cite{biggio2018}. In the context of VLMs, evasion attacks on visual inputs are particularly relevant, as they can silently alter the model's textual output without any visible change to the image.

\paragraph{Textual adversarial attacks.}
Zou et al.~\cite{zou2023} showed that gradient-based optimization can generate adversarial suffixes that bypass safety alignment in LLMs, with transferability across models including closed-source systems. While focused on text-only attacks, this work highlights a fundamental vulnerability of aligned models that extends to multimodal settings.

\paragraph{Visual adversarial attacks on VLMs.}
Several works have demonstrated that adversarial perturbations applied to images can manipulate VLM outputs. Qi et al.~\cite{qi2024} showed that a single optimized adversarial image can universally jailbreak aligned VLMs, establishing a connection between classical adversarial examples and multimodal alignment vulnerabilities. Schlarmann and Hein~\cite{schlarmann2023} studied both targeted and untargeted attacks against models such as OpenFlamingo, but their approach performs gradient-based optimization through the entire multimodal architecture, making it computationally expensive. In the black-box setting, Zhao et al.~\cite{zhao2023} proposed AttackVLM, which crafts adversarial examples using surrogate models such as CLIP and BLIP and transfers them to victim VLMs. Zhang et al.~\cite{zhang2024} extended these attacks to autonomous driving, demonstrating trajectory deviations in 70\% of physical trials, though their method also optimizes through the full multimodal pipeline.

\begin{table}
\caption{Comparison of VLM and vision encoder parameter counts.}\label{tab1}
\centering
\normalsize
\setlength{\arrayrulewidth}{0.3pt}
\begin{tabular}{lcccc}
\hline
\textbf{VLM}  & \textbf{Qwen} & \textbf{Granite} & \textbf{FastVLM} & \textbf{Phi} \\
\hline
VLM Params & 3.7B & 3B & 7.8B & 4.1B \\
\hline
VT Params  & 670M & 442M & 125M & 424M \\
\hline
Ratio      & 18\% & 14\% & 1.6\% & 10\% \\
\hline
\end{tabular}
\end{table}

\paragraph{Our approach.}
In contrast to prior work that requires backpropagation through the entire VLM, we optimize adversarial perturbations exclusively through the vision encoder. As shown in Table~\ref{tab1}, the vision encoder represents only 1.6\% to 18\% of the total model parameters, and this ratio decreases further for larger LLM backbones (e.g., 0.9\% for Qwen 72B which shares the same vision encoder as Qwen 3B). This design significantly reduces the computational and memory footprint of the attack while maintaining strong attack effectiveness, as further validated experimentally in Section 4.2.

\section{Problem formulation and attack design}

\subsection{Threat model}

We characterize the threat model along two dimensions: the attacker's goal and the attacker's knowledge.

\subsubsection{Attacker's goal:} In real-world scenarios, a malicious attacker could exploit VLM vulnerabilities to spread disinformation, bias model outputs, or propagate misleading information. We consider two attack settings:

\begin{itemize}
    \item \textbf{Untargeted attack:} the attacker aims to produce an adversarial image whose VLM-generated description no longer reflects the main semantic content of the original image.
    \item \textbf{Targeted attack:} the attacker starts from a source image and a target image, and aims to generate an adversarial image that remains visually identical to the source while causing the VLM to produce a description corresponding to the target image.
\end{itemize}

The detailed evaluation protocol, including the prompting strategy and success criteria, is described in Section~4.1.

\subsubsection{Attacker's knowledge:} We consider a white-box threat model in which the attacker has full access to the vision encoder of the VLM. This assumption is realistic in practice, as many competitive VLMs are publicly available as open-source systems, including Granite-Vision, Qwen-VL, Phi-Vision, and FastVLM.

\subsection{Design of our solution}

\vspace{-1cm}

\begin{figure}
\includegraphics[width=\textwidth]{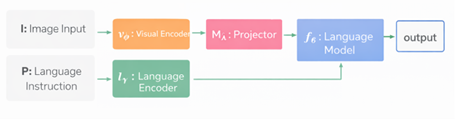}
\vspace{-1cm}
\caption{General architecture of a Vision--Language Model combining a visual encoder and a language model to generate textual outputs from image and text inputs.}\label{fig3}
\end{figure}

\vspace{-0.3cm}

\noindent Figure~\ref{fig3} illustrates the general architecture of a Vision--Language Model (VLM). The model receives two inputs: an image input $I$ and a prompt $P$. The image is first processed by the visual encoder $v_\phi$, which extracts visual features and converts them into visual embeddings. These embeddings are then mapped to the language embedding space through a projection module $M_\lambda$. The projected visual representations are subsequently combined with the textual instruction processed by the language encoder $l_\gamma$. Finally, the language model $f_\theta$ generates the textual output conditioned on both visual and textual information. \\

Let us denote by
\begin{equation*}
h = v_\phi(I)
\end{equation*}
the visual embedding produced by the vision encoder for an image $I$. \\

In our work, we exploit the differentiable nature of the visual encoder to generate adversarial perturbations on the image input. Instead of performing backpropagation through the entire multimodal model, which would involve both the visual and language components, we restrict the optimization process to the vision encoder alone. More specifically, the adversarial perturbations are computed by minimizing a loss defined on the image embeddings produced by the vision encoder. This approach significantly reduces the computational complexity of the attack while still allowing effective manipulation of the model's visual representation. \\

For targeted attacks, the attacker aims to force the VLM to interpret the adversarial image as a specific target image. The attack starts from a source image $I_{source}$ and a target image $I_{target}$. \\

We denote the embedding of the target image as
\begin{equation*}
h_{target} = v_\phi(I_{target})
\end{equation*}
and the embedding of the adversarial image as
\begin{equation*}
h_{adv} = v_\phi(I_{adv})
\end{equation*}

The loss function is defined as:
\begin{equation*}
\mathcal{L}_{targeted} = D(h_{adv}, h_{target})
\end{equation*}
where $D(\cdot,\cdot)$ denotes the Mean Squared Error (MSE) between embeddings. By minimizing this loss, the optimization process progressively brings the adversarial image embedding closer to that of the target image, gradually pushing the VLM to generate a description consistent with the target rather than the original. \\

For untargeted attacks, no specific target image is available. Instead, we initialize the optimization from a slightly noisy version of the original image, obtained by adding a Gaussian perturbation with a very small scale factor ($\sigma = 10^{-6}$) to the image tensor, which is then clipped to the valid pixel range $[0, 1]$. \\

Let $h_{source} = v_\phi(I_{source})$ be the embedding of the source image. The adversarial perturbation is then iteratively updated to maximize the distance between the adversarial and source embeddings:
\begin{equation*}
\mathcal{L}_{untargeted} = -D(h_{adv}, h_{source})
\end{equation*}

This forces the adversarial image representation to diverge from the original embedding, ultimately causing the VLM to misinterpret the image content. \\

An additional challenge arises from the fact that different VLM models rely on different vision encoder architectures. While most modern models use transformer-based visual encoders such as Vision Transformers (ViT), the internal structure and layer access can vary significantly across implementations. Some models expose intermediate feature layers, while others only provide final embeddings through specific projection modules. As a result, the implementation of our attack must be adapted to each model in order to correctly access the visual representations used during optimization. Despite these architectural differences, the core principle remains the same: adversarial perturbations are generated by optimizing the visual embeddings produced by the vision encoder. 

\begin{algorithm}
\caption{Vision-Encoder Targeted Adversarial Attack}\label{alg1}
\begin{algorithmic}[1]
\REQUIRE benign image $I_b$, target image $I_t$, vision encoder $v_\phi$, number of iterations $N$, step size $\alpha$, perturbation limit $\varepsilon$
\ENSURE adversarial image $I_{adv}$
\STATE $I_{adv} \leftarrow I_b$
\STATE $\text{target\_embedding} \leftarrow v_\phi(I_t)$
\FOR{$k = 1$ \TO $N$}
    \STATE $\text{adv\_embedding} \leftarrow v_\phi(I_{adv})$
    \STATE $\text{loss} \leftarrow \text{MSE}(\text{adv\_embedding}, \text{target\_embedding})$
    \STATE $\text{gradients} \leftarrow \nabla_{I_{adv}}(\text{loss})$
    \STATE $I_{adv} \leftarrow I_{adv} - \alpha \cdot \text{gradients}$
    \STATE $I_{adv} \leftarrow \text{Clip}(I_{adv}, I_b - \varepsilon, I_b + \varepsilon)$
\ENDFOR
\RETURN $I_{adv}$
\end{algorithmic}
\end{algorithm}

\noindent A pseudocode for the targeted attack procedure is presented in Algorithm~\ref{alg1}. The untargeted attack follows the same general optimization procedure but differs in two key aspects: it requires no target image and uses a negated MSE loss to maximize the divergence between embeddings rather than minimizing it (see Algorithm~\ref{alg2} in Appendix). In both cases, the optimization is performed directly on the image input while backpropagating gradients only through the vision encoder. This allows efficient generation of adversarial images while avoiding the computational cost of optimizing through the entire multimodal architecture.

\section{Experiment}

\subsection{Methodology and settings}

To perform our experiments, we used a subset of the ImageNet dataset, which contains a large collection of images covering a wide variety of objects and scenes. This diversity makes it particularly suitable for evaluating the robustness of vision--language models against adversarial perturbations. \\

To demonstrate the effectiveness of our targeted and untargeted attacks, we designed experiments under a general and realistic attack setting. For the targeted attacks, we randomly selected 1000 images from the dataset to serve as source images. Each source image was then randomly paired with another image from the dataset to serve as the target image. This procedure resulted in 1000 image pairs $(I_{source}, I_{target})$. Because the images were randomly selected, most pairs are semantically unrelated, making the attack task particularly challenging. \\

Adversarial perturbations were generated using I-FGSM (Iterative Fast Gradient Sign Method), an iterative extension of the FGSM attack introduced by Goodfellow et al.~\cite{goodfellow2015}, applied exclusively through the vision encoder as described in Section~3. The optimization was run for 50 iterations for all experiments. \\

For the untargeted attacks, we used the same set of original images but without pairing them with target images. Instead, we initialized the adversarial optimization from a slightly noisy version of the original image, obtained by adding a Gaussian perturbation. The optimization then iteratively modifies the image in order to disrupt the semantic interpretation produced by the vision--language models. \\

To evaluate whether an attack was successful, we adopted an LLM-as-a-Judge evaluation method. After generating an adversarial image, we queried each VLM to produce a one-sentence description of the image --- consistent with the evaluation protocol defined in Section~3.1. The generated descriptions were then compared using a separate language model acting as a semantic evaluator. Specifically, we used Granite-4.0-micro as the judging model. The judge receives two textual descriptions and determines whether they describe the same semantic content or not. For targeted attacks, the judge verifies whether the description generated from the adversarial image is semantically consistent with the target image. For untargeted attacks, the judge verifies whether the description generated from the adversarial image is semantically different from the description of the original image. The prompts used for this evaluation procedure are shown in Figure~\ref{fig4} in Appendix. \\

We conducted our experiments on four open-source vision--language models: Qwen2.5-VL-3B-Instruct, Granite-Vision-3.2-2B, FastVLM-7B, and Phi-3.5-Vision-Instruct. These models were selected because they represent competitive open-source VLM architectures while remaining computationally manageable for large-scale adversarial experiments. Although these models may not reach the performance level of the largest proprietary systems, they still demonstrate strong multimodal capabilities and therefore provide a realistic testbed for evaluating adversarial robustness. Moreover, previous studies have shown that larger machine learning models do not necessarily exhibit greater robustness to adversarial perturbations~\cite{szegedy2013}. \\

All experiments were conducted on a server equipped with a GPU cluster. We used a single NVIDIA H100 GPU with 80~GB of VRAM, which was sufficient to run the adversarial optimization procedure across all evaluated models.

\subsection{Discussion of the results}

\noindent An example of this attack on Granite-Vision-3.2-2B with $\varepsilon=0.05$ is shown in Figure~\ref{fig5} in Appendix. The adversarial image is visually indistinguishable from the original, yet the model's output changes from ``a military tank'' to ``an ambulance''. To evaluate the effectiveness of our attack, we measured the attack success rate for each model and perturbation budget $\epsilon$. The reported values correspond to the mean success rate over four independent runs, and the $\pm$ values represent the standard deviation ($\Delta$) across these runs.

\subsubsection{Targeted attacks:} 
\leavevmode\newline

\begin{figure}
\includegraphics[width=\textwidth]{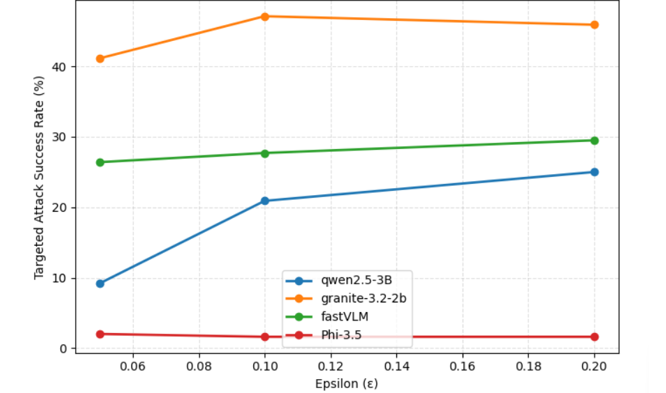}
\vspace{-0.6cm}
\caption{Targeted attack success rate as a function of the perturbation budget $\epsilon$.}\label{fig6}
\end{figure}

\noindent Table~\ref{tab2} and Figure~\ref{fig6} present the success rates for targeted attacks. Overall, as expected, increasing the perturbation budget $\varepsilon$ generally improves the attack success rate. However, the vulnerability varies significantly across models.

It is important to recall that source-target image pairs were chosen randomly, meaning that most pairs are semantically unrelated. The attack may for instance attempt to make a model describe a dog in a meadow as a bedside lamp in a bedroom. This makes the targeted attack setting particularly challenging, as the adversarial perturbation must bridge a large semantic gap while remaining imperceptible to the human eye. The success rates reported here should therefore be interpreted in this context.

\begin{table}[h]
\centering
\caption{Targeted attack success rate (\%) across VLM models for different perturbation budgets $\epsilon$. Mean $\pm$ standard deviation over four runs.}\label{tab2}
\resizebox{0.85\textwidth}{!}{%
\begin{tabular}{|c|c|c|c|c|}
\hline
\textbf{epsilon} & \textbf{qwen2.5-3B} & \textbf{granite-3.2-2b} & \textbf{FastVLM} & \textbf{Phi-3.5}\\
\hline
\textbf{0.05} & $9.2 \pm 0.8$ & $41.15 \pm 1.9$ & $26.4 \pm 1.3$ & $2.0 \pm 0.4$\\
\textbf{0.10} & $20.9 \pm 1.4$ & $47.12 \pm 2.1$ & $27.7 \pm 1.6$ & $1.6 \pm 0.3$\\
\textbf{0.20} & $25.0 \pm 1.7$ & $45.91 \pm 2.4$ & $29.5 \pm 1.8$ & $1.6 \pm 0.3$\\
\hline
\end{tabular}%
}
\end{table}

\noindent Despite this difficulty, Granite-Vision-3.2-2B reaches success rates above 45\% for larger perturbations, which is a remarkably high figure given the semantic distance between random image pairs. This suggests that the visual representations learned by this model are particularly susceptible to embedding-space manipulation. FastVLM shows moderate vulnerability with success rates between 26\% and 29\%, indicating that while the model is not easily fooled, a non-negligible fraction of attacks succeed even in unfavorable semantic conditions. Qwen2.5-VL-3B demonstrates greater resilience at small perturbation budgets but becomes increasingly vulnerable as $\varepsilon$ grows, suggesting that its robustness has clear limits when the attacker is allowed slightly larger perturbations. In contrast, Phi-3.5-Vision remains highly robust across all perturbation levels, with success rates consistently around 2\%, making it an outlier among the evaluated models and a potentially interesting architecture to study from a robustness perspective. This variability in robustness across architectures is consistent with prior findings suggesting that there exist fundamental trade-offs between standard accuracy and adversarial robustness~\cite{tsipras2019}, and that different architectural choices can lead to significantly different levels of vulnerability.

\subsubsection{Untargeted attacks:}
\leavevmode\newline

\begin{figure}
\includegraphics[width=\textwidth]{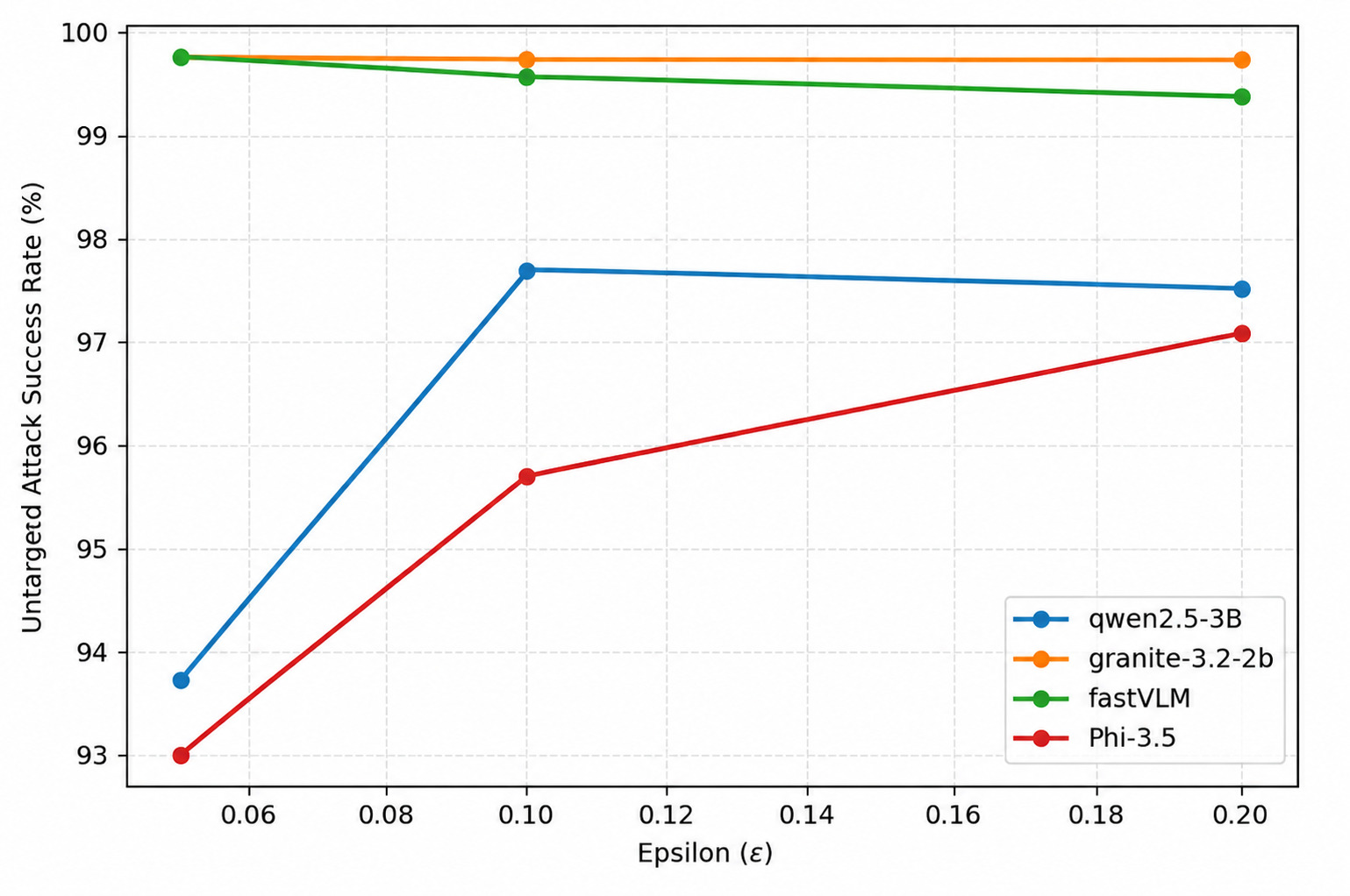}
\vspace{-0.6cm}
\caption{Untargeted attack success rate as a function of the perturbation budget $\epsilon$.}\label{fig7}
\end{figure}

\noindent Table~\ref{tab3} and Figure~\ref{fig7} show the results for untargeted attacks. In this setting, attacks are extremely effective across all models. Even for small perturbations ($\varepsilon = 0.05$), the success rate already exceeds 93\% for every model, which indicates that VLMs are highly sensitive to even minimal visual perturbations when no specific target is imposed.

\begin{table}
\centering
\caption{Untargeted attack success rate (\%) across VLM models for different perturbation budgets $\epsilon$. Mean $\pm$ standard deviation over four runs.}\label{tab3}
\resizebox{0.85\textwidth}{!}{%
\begin{tabular}{|c|c|c|c|c|}
\hline
\textbf{epsilon} & \textbf{qwen2.5-3B} & \textbf{granite-3.2-2b} & \textbf{FastVLM} & \textbf{Phi-3.5}\\
\hline
\textbf{0.05} & $93.7 \pm 1.1$ & $99.79 \pm 0.08$ & $99.8 \pm 0.07$ & $93.0 \pm 1.3$\\
\textbf{0.10} & $97.7 \pm 0.9$ & $99.78 \pm 0.07$ & $99.6 \pm 0.09$ & $95.7 \pm 1.0$\\
\textbf{0.20} & $97.5 \pm 1.0$ & $99.77 \pm 0.06$ & $99.4 \pm 0.12$ & $97.1 \pm 0.8$\\
\hline
\end{tabular}%
}
\end{table}

Granite-Vision and FastVLM show the highest vulnerability, with success rates close to 99--100\% across all perturbation levels. This near-perfect attack success suggests that the visual embeddings produced by these encoders are particularly unstable. Small input perturbations are sufficient to push their representations far from the original, effectively erasing the semantic content of the image. Qwen2.5-VL and Phi-3.5-Vision also exhibit very high success rates above 97\%, confirming that no evaluated model offers meaningful resistance to untargeted perturbations.

Interestingly, increasing the perturbation budget $\varepsilon$ beyond 0.05 yields only marginal improvements across all models, suggesting that the attack already saturates at low perturbation levels. This stands in sharp contrast with the targeted attack setting, where larger $\varepsilon$ values led to more noticeable gains. This observation implies that disrupting semantic coherence is a fundamentally easier objective than steering the model toward a specific description, and that the vision encoders of current VLMs are inherently fragile to even imperceptible input modifications. \\

\vspace{-0.8cm}

\vspace{-0.6cm}

\noindent Overall, these results indicate that disrupting the semantic interpretation of an image is significantly easier than forcing a model to produce a specific targeted description. This raises serious concerns about the deployment of VLMs in safety-critical applications, where the integrity of visual inputs cannot be guaranteed, such as autonomous driving, medical imaging, or content moderation systems. In particular, Zhang et al. (2024)~\cite{zhang2024} confirmed these risks by showing that adversarial perturbations on autonomous driving VLMs can cause real vehicles to deviate from their intended trajectories.

\subsubsection{Computational cost and resource usage:}
\leavevmode\newline

\noindent Beyond attack effectiveness, we evaluated the resource footprint of the proposed vision-encoder-only optimization on Qwen2.5-VL and Granite-Vision, the two models with the highest vision-encoder-to-VLM parameter ratios. On an NVIDIA H100 80 GB, restricting backpropagation to the vision encoder reduces VRAM usage from approximately 44--47\% for the full VLM to 13--16\%, while also substantially reducing GPU utilization. Most importantly, this reduction translates into a much shorter attack generation time: on Qwen2.5-VL, generating an adversarial example with the full VLM required more than 20 minutes without success, whereas the vision-encoder-only formulation required approximately 100 seconds. These measurements provide empirical evidence that the proposed approach is not only effective, but also substantially more resource-efficient than full-pipeline optimization. Detailed GPU compute and VRAM usage measurements for both models are provided in Figures~\ref{fig:granite_gpu} and~\ref{fig:qwen_gpu} in the Appendix.

\section{Conclusion and future work}

In this work, we investigated the robustness of Vision Language Models to adversarial visual perturbations. We introduced an efficient gradient-based attack that operates exclusively on the vision encoder, allowing adversarial examples to be generated without backpropagating through the entire VLM. Experimental measurements further show that this formulation substantially reduces VRAM usage, GPU utilization, and attack generation time while maintaining strong effectiveness across multiple models. Our findings highlight that current VLMs remain vulnerable to visual adversarial manipulation despite their strong multimodal capabilities.

Future work could explore several directions. First, an important direction would be to investigate the transferability of adversarial perturbations across different VLM architectures. Second, investigating relevant defense mechanisms.

%
%
%

\newpage
\section*{Appendix}
\vspace{-0.5cm}

\begin{algorithm}
\caption{Vision-Encoder Untargeted Adversarial Attack}\label{alg2}
\begin{algorithmic}[1]
\REQUIRE benign image $I_b$, vision encoder $v_\phi$, number of iterations $N$, step size $\alpha$, perturbation limit $\varepsilon$
\ENSURE adversarial image $I_{adv}$
\STATE $\text{source\_embedding} \leftarrow v_\phi(I_b)$
\STATE $I_{adv} \leftarrow I_b + 10^{-6} \cdot \text{randn\_like}(I_b)$
\STATE $I_{adv} \leftarrow \text{Clip}(I_{adv}, 0, 1)$
\FOR{$k = 1$ \TO $N$}
    \STATE $\text{adv\_embedding} \leftarrow v_\phi(I_{adv})$
    \STATE $\text{loss} \leftarrow -\text{MSE}(\text{adv\_embedding}, \text{source\_embedding})$
    \STATE $\text{gradients} \leftarrow \nabla_{I_{adv}}(\text{loss})$
    \STATE $I_{adv} \leftarrow I_{adv} - \alpha \cdot \text{gradients}$
    \STATE $I_{adv} \leftarrow \text{Clip}(I_{adv}, I_b - \varepsilon, I_b + \varepsilon)$
\ENDFOR
\RETURN $I_{adv}$
\end{algorithmic}
\end{algorithm}

\begin{figure}[H]
\includegraphics[width=\textwidth]{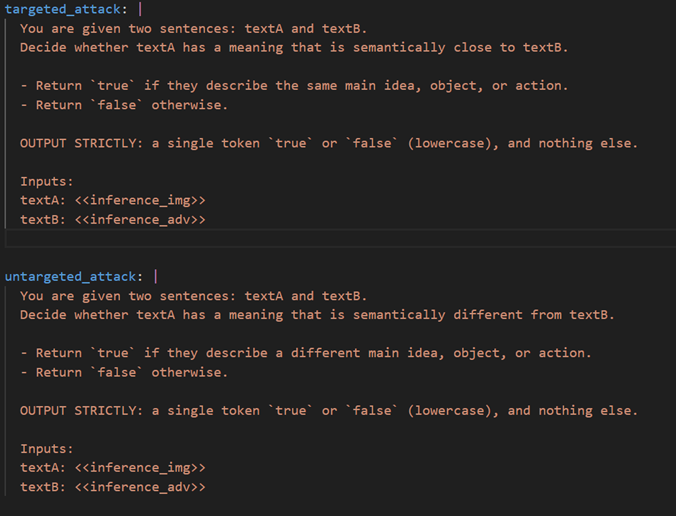}
\caption{Prompts used for the LLM-as-a-Judge evaluation to assess semantic similarity or difference between generated image descriptions.}\label{fig4}
\end{figure}

\begin{figure}[H]
\includegraphics[width=\textwidth]{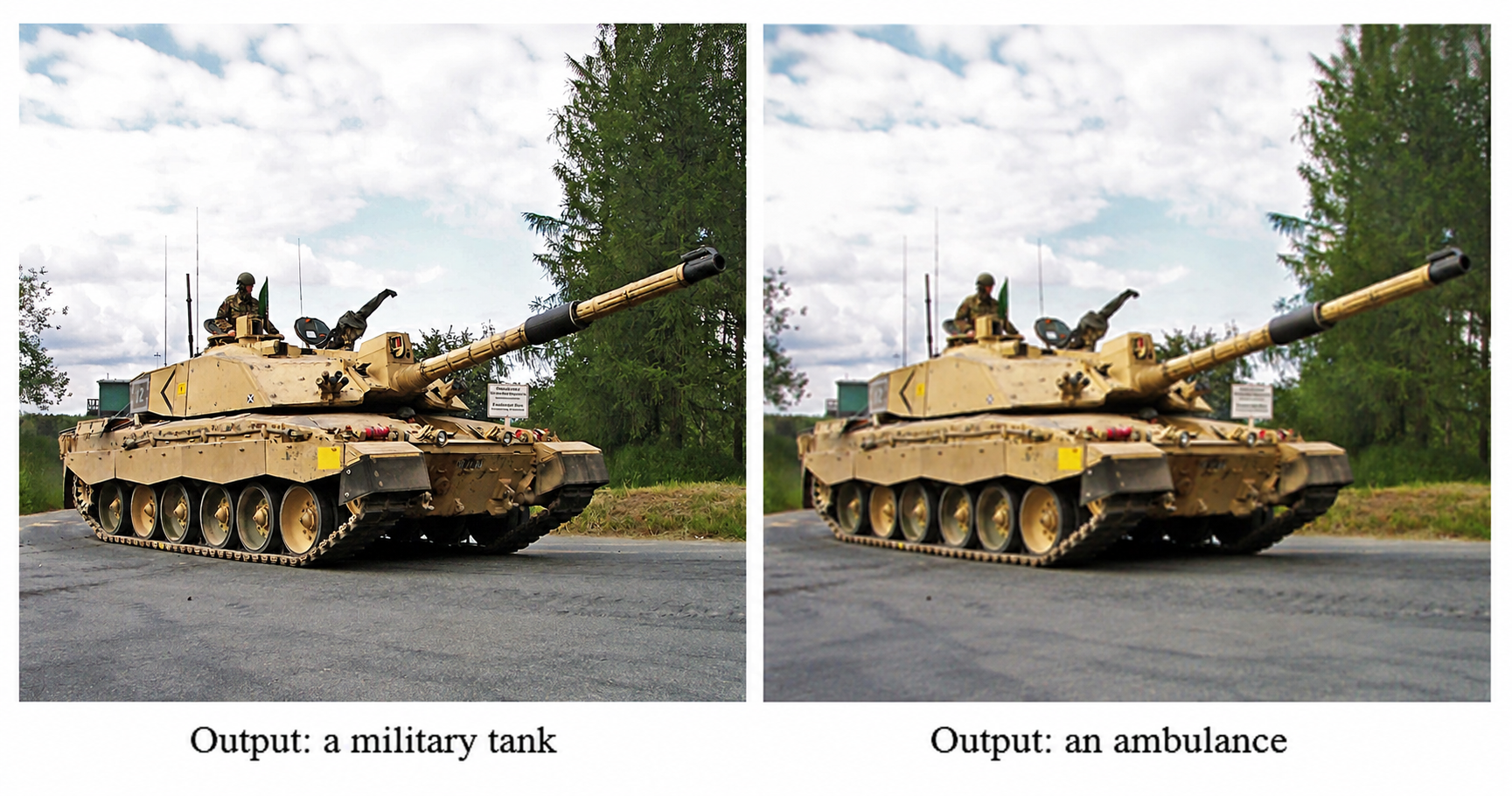}
\caption{Generated captions on source (left) and adversarial (right) images. We perform a targeted attack on the caption output with $\varepsilon=0.05$ on Granite-Vision-3.2-2B. The perturbations are hardly visible and would not be noticed by a user.}\label{fig5}
\end{figure}

\begin{figure}[H]
    \centering
    \begin{minipage}{0.82\textwidth}
        \centering
        \includegraphics[width=\textwidth]{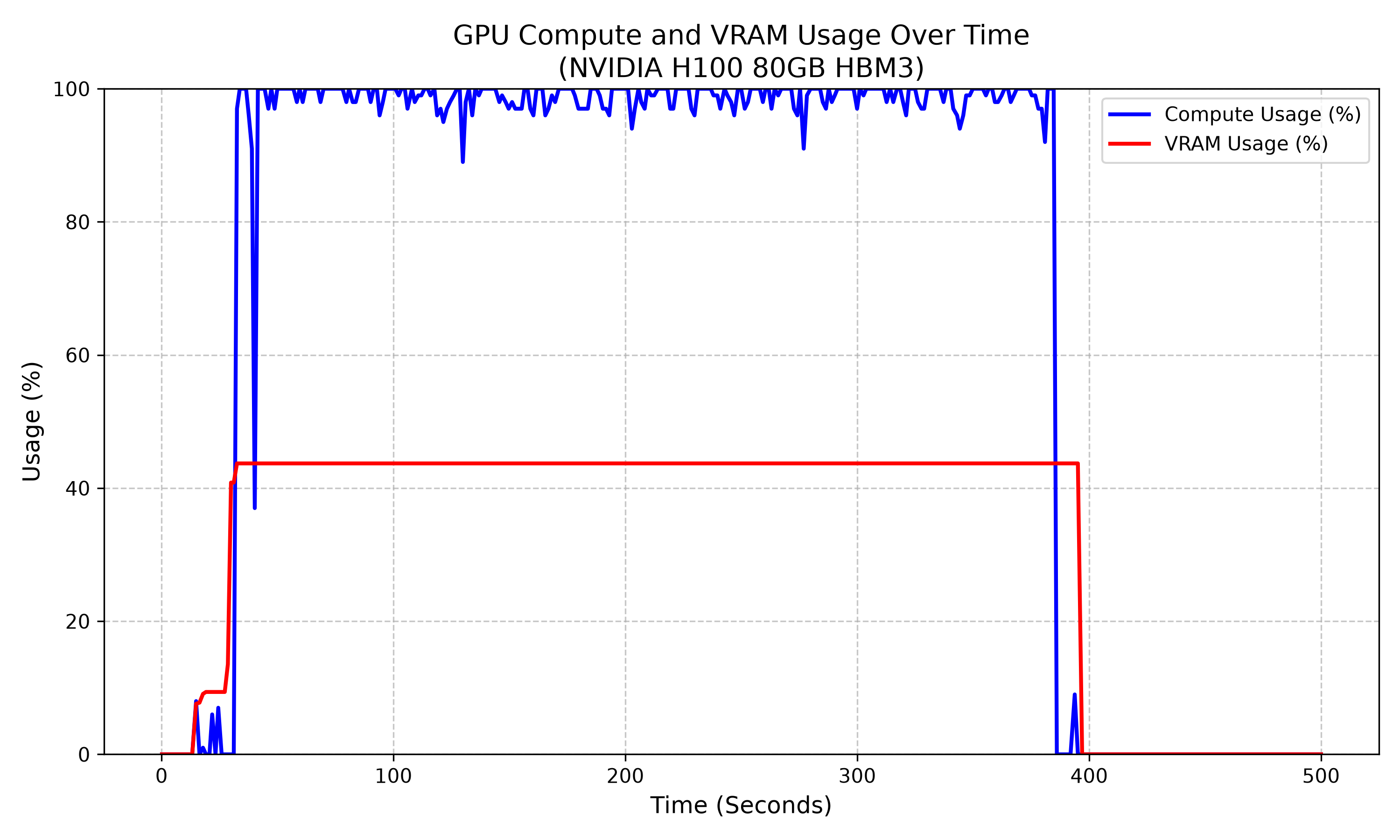}
        \vspace{0.15cm}
        \includegraphics[width=\textwidth]{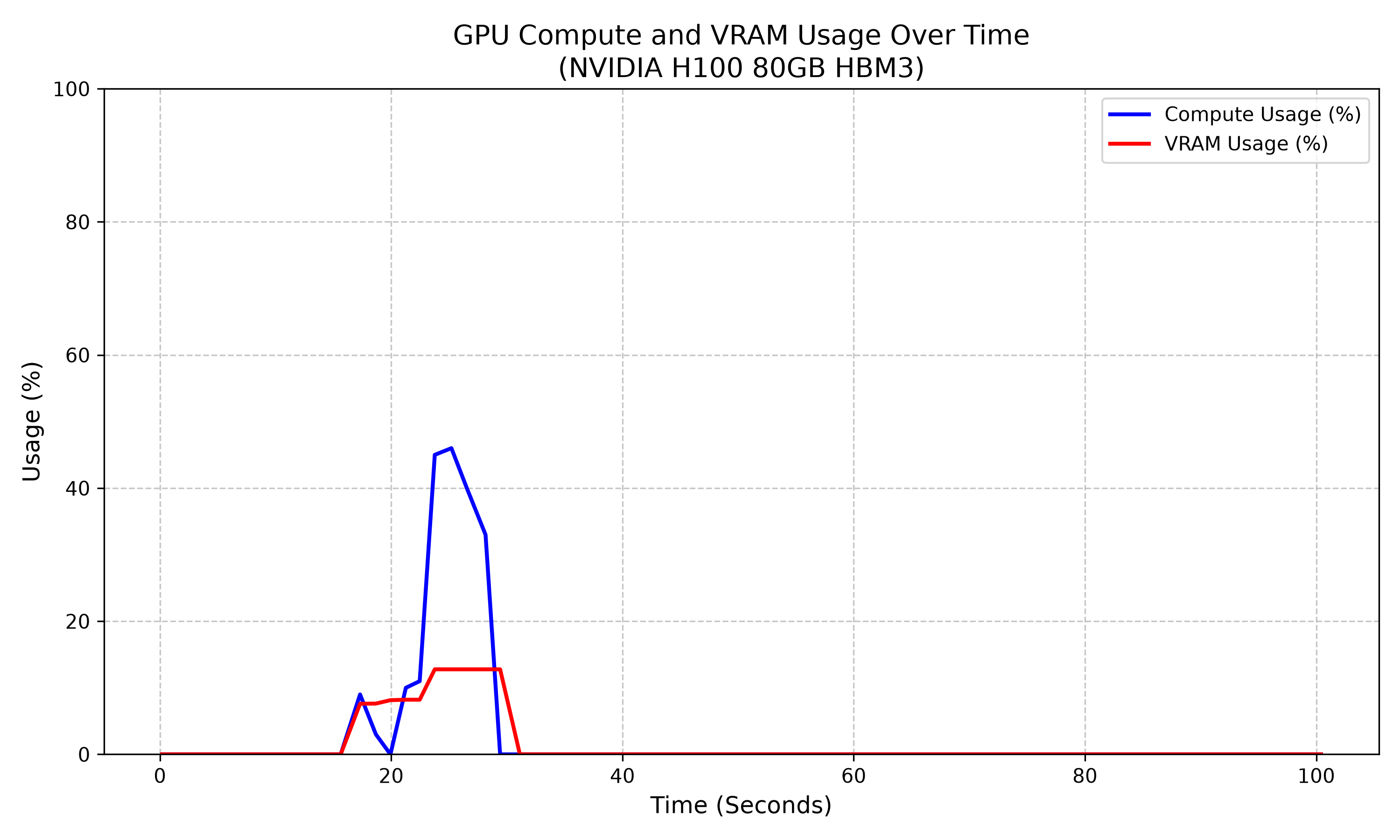}
    \end{minipage}
    \caption{GPU compute and VRAM usage over time for Granite-Vision-3.2-2B on an NVIDIA H100 80 GB. Top: optimization through the full VLM. Bottom: optimization restricted to the vision encoder. Encoder-only optimization substantially reduces both VRAM usage and GPU utilization.}
    \label{fig:granite_gpu}
\end{figure}

\begin{figure}[H]
    \centering
    \begin{minipage}{0.82\textwidth}
        \centering
        \includegraphics[width=\textwidth]{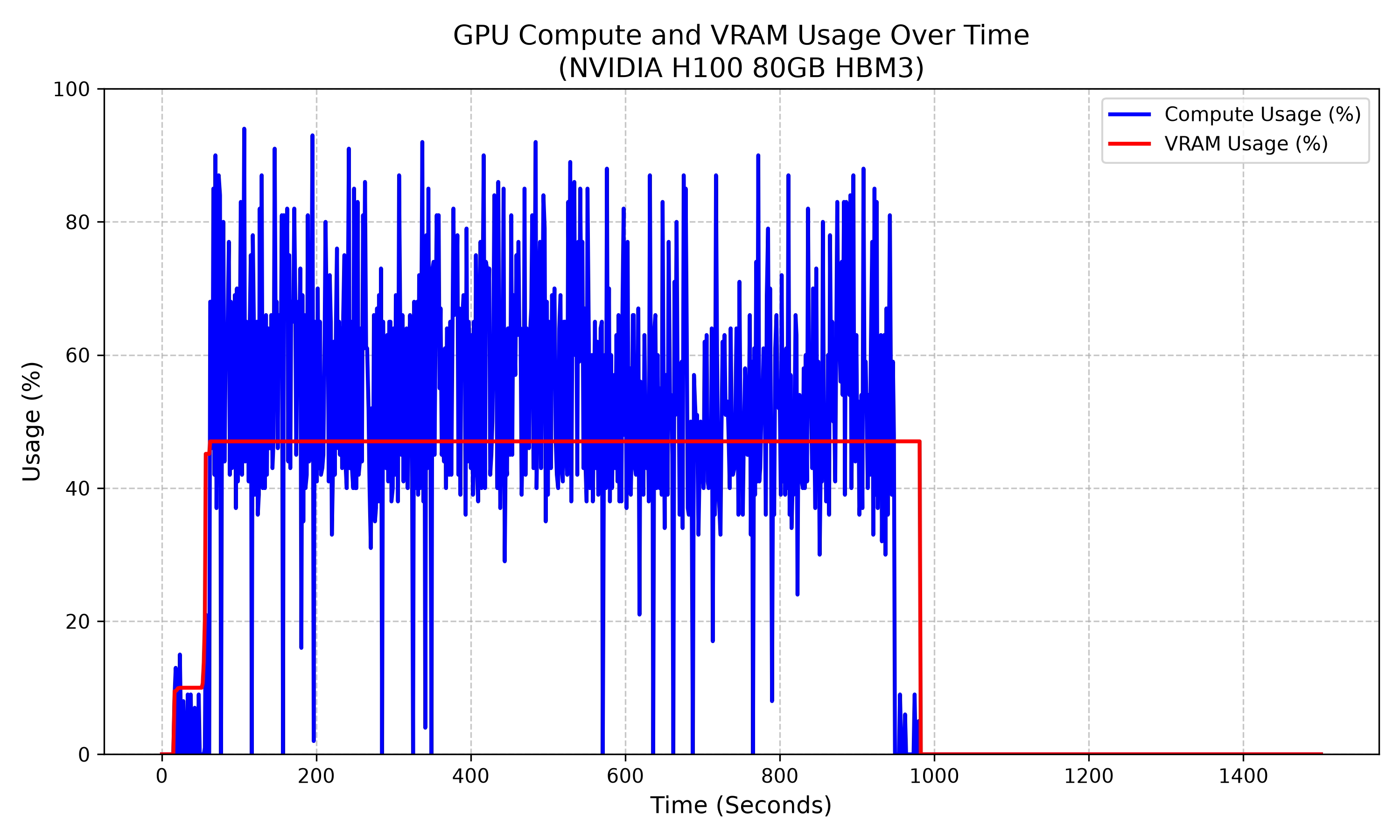}
        \vspace{0.15cm}
        \includegraphics[width=\textwidth]{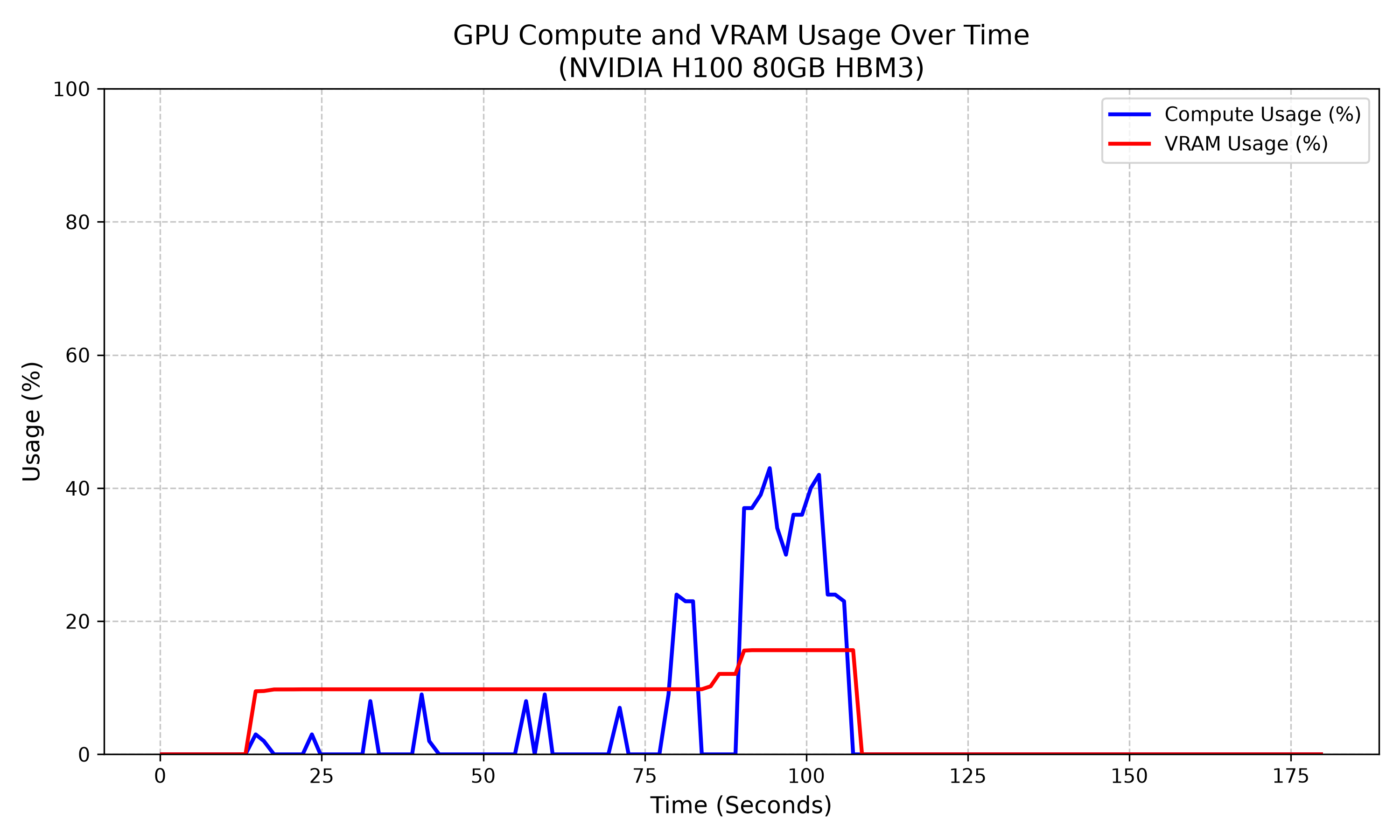}
    \end{minipage}
    \caption{GPU compute and VRAM usage over time for Qwen2.5-VL-3B on an NVIDIA H100 80 GB. Top: optimization through the full VLM. Bottom: optimization restricted to the vision encoder. The full-VLM experiment may require more than 20 minutes without producing a successful adversarial example, whereas the encoder-only approach completes in approximately 100 seconds.}
    \label{fig:qwen_gpu}
\end{figure}

\end{document}